# An embarrassingly simple approach to neural multiple instance classification

Amina Asif[a] and Fayyaz ul Amir Afsar Minhas[a, *]

[a] *PIEAS Data Science Lab, Department of Computer and Information Sciences, Pakistan Institute of Engineering and Applied Sciences (PIEAS), PO Nilore, Islamabad, Pakistan.*

## ABSTRACT

Multiple Instance Learning (MIL) is a weak supervision learning paradigm that allows modeling of machine learning problems in which labels are available only for groups of examples called bags. A positive bag may contain one or more positive examples but it is not known which examples in the bag are positive. All examples in a negative bag belong to the negative class. Such problems arise frequently in fields of computer vision, medical image processing and bioinformatics. Many neural network based solutions have been proposed in the literature for MIL, however, almost all of them rely on introducing specialized blocks and connectivity in the architectures. In this paper, we present a novel and effective approach to Multiple Instance Learning in neural networks. Instead of making changes to the architectures, we propose a simple bag-level ranking loss function that allows Multiple Instance Classification in any neural architecture. We have demonstrated the effectiveness of our proposed method for popular MIL benchmark datasets. In addition, we have tested the performance of our method in convolutional neural networks used to model an MIL problem derived from the well-known MNIST dataset. Results have shown that despite being simpler, our proposed scheme is comparable or better than existing methods in the literature in practical scenarios. Python code files for all the experiments can be found at https://github.com/amina01/ESMIL.

## 1. Introduction

Supervised machine learning methods work by training a model over a labeled set of training examples and then deploying it for testing after performance evaluation [1]. Conventional supervised methods require accurately labeled examples for training. Any noise or ambiguity in the labels can affect learning and, hence, the test performance of a classifier [2]. Scenarios involving labeling ambiguities arise quite often in machine learning problems and therefore, specialized methods are required that can handle such situations. One such weak supervision paradigm [3], known as Multiple Instance Learning (MIL), is aimed to model problems in which training labels are not available for individual examples [4], [5]. Rather, labels are associated with groups of examples called *bags*. Specifically, a bag with a positive label implies that at least one of the constituent examples is positive. However, it is not known which examples in the bag belong to the positive class. On the other hand, all examples in a negatively labeled bag are negative. The concept is illustrated in Figure 1. The task in Multiple Instance Classification is to learn a model that,
given training data in the form of bags, can classify test data both in the form of individual examples and bags.

Multiple Instance Learning has a number of applications in areas of computer vision, bioinformatics and medical image processing [6]–[9]. For instance, consider the development of a machine learning based object detection or tracking system for which the training data consists of annotated frames in videos, i.e., a frame is labeled positive if it contains the object of interest and negative otherwise, but the exact location of the object is not known. The lack of patch-based labeling of frames in the videos to be used for training makes it a Multiple Instance Learning problem. Multiple Instance Learning has successfully been used for modeling such visual tracking problems in [6], [8], [9] . MIL is also widely applicable in the domain of bioinformatics such as protein function annotation. Proteins are macromolecules composed of a sequence of amino acids that perform most of the functions in living organisms [10]. In machine learning based protein function annotation, the objective is to develop a machine learning system that can predict whether a given protein performs a particular function (e.g., Amyloid formation, binding, etc.) or not given its sequence. The whole of a protein may not be responsible for a particular function, but training annotations are typically only available for the whole protein sequence. As a consequence of such labeling ambiguities, conventional machine learning classification approaches that require instance level labels are not suitable for these problems. Multiple Instance Learning has been used effectively for modeling such problems, e.g., prediction of Calmodulin binding sites in proteins [11], [12], studying protein-protein interactions [13], functional annotation of proteins [14], prediction of protein-ligand binding affinities [15].

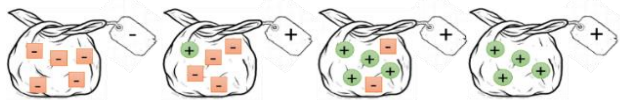

**Figure 1- Illustration of concept of bags. A bag is labeled positive if at least one of the constituent examples is positive. A bag is labeled as negative if all the examples belong to the negative class**

* Corresponding author. +92-51-2207381 to 85, Extension: 3164; Fax: +92-51-9248600; e-mail:  *afsar@pieas.edu.pk*; *fayyazafsar@gmail.com*

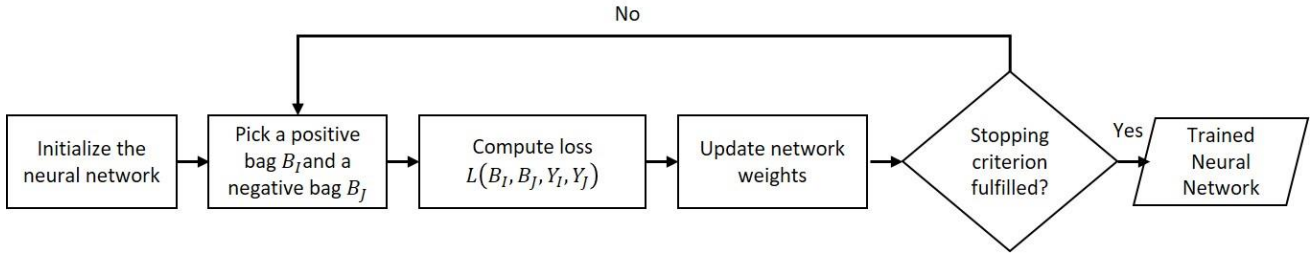

**Figure 2- Illustration of training process of a neural network using the proposed loss.**

There are several techniques in the literature for Multiple Instance Learning. The concept of Multiple Instance Learning and its solution using parallel axis rectangles was first proposed by Dietterich et al. for drug activity prediction [5]. Dooly et al. proposed extension of k Nearest Neighbor (k-NN) and Diverse Density (DD) for Multiple Instance Learning with real valued targets [16]. EM-DD, a solution combining Expectation Maximization (EM) and Diverse Density for MIL, was presented by by Zhang et al. in [17]. Gärtner et al proposed specialized kernels using which methods such as Support Vector Machines (SVMs) could be used for Multiple Instance Learning [18]. Andrews et al. proposed two heuristic solutions to SVMs for MIL: one performing bag level classification (MI-SVM), the other instance level classification (mi-SVM) [19]. Another solution for MIL, that mapped bags to graphs and defined graph kernels, called mi-Graph was proposed in [20]. Wei et al. proposed scalable MIL solutions for large datasets using two new mappings for representation of bags: one based on locally aggregated descriptors called miVLAD and the other using Fisher vector representation called miFV [21]. Other popular solutions include Multiple-Instance Learning via Embedded Instance Selection (MILES) [22], deterministic annealing for MIL [23], semi-supervised SVMs for MIL (MissSVM) [24], generalized dictionaries for MIL [25], MIL with manifold bags [26], MIL with randomized trees [27]. Apart from these, many neural network based solutions had also been proposed for Multiple Instance Learning [28]–[30]. With recent advances in deep learning, deep neural networks for Multiple Instance Learning such as Convolutional Neural Networks (CNNs) based MIL architectures have also been proposed [7], [31]–[33]. Wang et al. proposed specialized pooling layers and residual connections to perform MIL in neural networks [34]. Recently, an attention networks based approach for deep MIL was proposed by Ilse et al. [35]. Most of the neural networks based approaches rely on the use of specialized pooling layers and connectivity to perform Multiple Instance Learning.

In this paper, we present a simple yet effective method to perform Multiple Instance Learning in neural networks. We propose a novel ranking-like loss function that can be used to implement MIL without any specialized layers or connectivity. The proposed training scheme can be used with any architecture of choice. Experiments over different MIL datasets have proven the effectiveness of the proposed technique. In section 2, mathematical formulation and experimental setup employed for evaluation of the method have been presented. Results have been reported in section 3 followed by conclusion in section 4.

## 2. Methods

In this section, we present the mathematical formulation of the proposed method and experimental setup employed to evaluate its performance.

### 2.1. Mathematical Formulation

In a typical Multiple Instance Learning (MIL) scenario, we are given $N$ non-overlapping bags $B_1, B_2, B_3, \ldots, B_N$ that have been created from $n$ examples $x_1, x_2, x_3, \ldots, x_n$ and associated bag labels $Y_I \in \{+1, -1\}, I = 1 \ldots N$. The objective is to learn a mathematical function $f(B_I; \theta)$ parameterized by $\theta$ given training data in the form of bags, such that, it can classify unseen bags and examples. The parameters $\theta$ can be thought of as weights of a neural network. As mentioned earlier, a positive bag may consist of one or more positive examples whereas all examples in a negative bag belong to the negative class. However, it is not known which examples in a positive bag are positive. Typically, supervised binary classification models are built such that the classifier is forced to produce a score above (below) a certain threshold (usually zero) for positive (negative) data points. This is ensured using a classification loss function where a penalty is imposed if the predictor does not produce positive scores for positive examples and negative scores for negative examples. Instead of using a threshold-based classification loss, we propose a ranking-like loss function at the bag level that imposes a penalty when the classifier produces a higher score for a negatively labeled bag as compared to a positive bag. Thus, given positive and negative bags $B_I$ and $B_J$, with $Y_I = +1$ and $Y_J = -1$ respectively, the objective of MIL can be interpreted as enforcing the constraint $f(B_I) > f(B_J)$ for all such pairs of bags in the training data. Therefore, the hinge loss function for this problem can be written as:

$$l(B_I, B_J, Y_I, Y_J) = \max\{0, 1 - (Y_I - Y_J)(f(B_I; \theta) - f(B_J; \theta))\}.$$

The minimization of this loss function during training will ensure that positively labeled bags always score higher than negatively labeled bags. We can define the score of a bag as the highest score produced by any of its constituent examples, i.e., without introducing further notation:

$$f(B; \theta) = \max_{x_i \in B} f(x_i; \theta).$$

The goal during training is to minimize the above-mentioned loss over all possible pairs of positive and negative bags. Thus the empirical error minimization problem underlying the proposed MIL scheme can be written as:

$$\theta^* = argmin_\theta \sum_{I,J=1, J>I}^{N} \max\{0, 1 - (Y_I - Y_J)(f(B_I; \theta) - f(B_J; \theta))\}$$

This minimization ensures that the highest scoring example in a positive bag should always be ranked higher than the highest scoring example in a negative bag. This property makes the proposed scheme better at maximizing AUC-ROC as compared to simple classification based losses [36]. Furthermore, using the paired-comparison based loss improves the quality of learning from small data sets.

As an alternative to adding specialized complex layers to make MIL work for neural networks, we propose a simpler approach of using the above-mentioned loss function for training. As shown in Figure 2, the iterative algorithm randomly picks a pair of bags (one positive and one negative) in each iteration and computes the scores of both bags and the resulting loss which is then back-propagated to update the weights of a neural network using an optimization scheme.

## 2.2. Experimental Setup

In this section, we present details of the experiments performed to evaluate the performance of our method. Description of the datasets, neural network architectures and evaluation metrics is presented in the following sections. Python code files for all the experiments can be found at https://github.com/amina01/ESMIL.

### 2.2.1. Datasets

We present the details of the datasets used for the performance evaluation of our method as follows.

*Benchmark Datasets*

We have performed evaluation of our method on five MIL benchmark datasets: MUSK-1, MUSK-2, Fox, Tiger and Elephant [5], [19].

MUSK-1 and MUSK-2 have been taken from the University of California, Irvine (UCI) repository of machine learning datasets [5]. MUSK-1/2 are drug activity prediction datasets. The task is to predict whether a molecule possesses musky nature or not [37]. A molecule may exist in multiple conformations but not all shapes are musky. A molecule is labeled positive if one or more of its conformations show muskiness and negative otherwise. Individual conformations are not labeled. All configurations of a molecule are grouped in a bag, that is, a bag represents a molecule and examples in a bag correspond to the possible conformations of that molecule. Each individual example is characterized using a 166 dimensional feature vector. MUSK-1 comprises of 47 positive and 45 negative bags with each bag containing 2 to 40 examples. MUSK-2 contains 39 positive and 63 negative bags. The smallest bag in MUSK-2 contains a single example while the largest has 1044 instances.

Fox, Tiger and Elephant datasets are subsets of Corel image retrieval dataset [19]. The task for each of the datasets is to identify if an image contains the animal the dataset is named after or not. Each image is divided into smaller segments. All the segments extracted from an image are grouped in a bag. That is, each bag represents an image and examples in a bag correspond to the patches extracted from that image. The examples are represented using color, texture and shape features for the image segments. Length of each feature vector is 230. A bag is labeled positive if the corresponding image contains the animal and negative otherwise. That is if one or more segments of the image contain the animal, the bag is given a positive label. Each of the three datasets comprise of 100 positive bags and 100 negative bags. Bags in these datasets contain 2 to 13 examples each.

*MNIST MIL Dataset*

To assess the effectiveness of our proposed scheme in classification models performing automatic feature extraction through Convolutional Neural Networks (CNNs) [38], we have replicated the MNIST [39] based experiment performed by Ilse et al. in [35]. MNIST is an image dataset comprising of handwritten digits from 0 to 9 of size 27×27 pixels. To test the performance of their proposed Attention Networks for MIL, Ilse et al. created a Multiple Instance dataset [35] derived from MNIST [39] for classification of 9 vs non-9 images in which images of numbers were grouped into bags. A bag is labeled positive if it contains one or more images of number 9 and negative otherwise. Note that the this is a hard classification problem as images of handwritten 9 are typically similar in structure to other numbers like 7 and 4. The number of samples per bag follow the Gaussian distribution with an average bag size of 10 instances per bag and a variance of 2.0. Performance of our method over varying number of training bags (50, 100, 150, 200, 300, 400, 500) has been studied. The size of test set has been fixed to 1000 bags. This evaluation protocol is the same as in [35] for a fair performance comparison.

### 2.2.2. Neural Network Architectures

For the benchmark datasets, we used two neural network architectures with the proposed loss function: a single layer neural network and another with one hidden layer. The first architecture corresponds to a single neuron with linear activation. The hidden layer in the second architecture contains the same number of neurons as in the input layer, i.e., equal to the example feature vector size (see figure 3 a, b). Tanh activation function is used for the hidden layer neurons and linear activation for the output layer neurons.

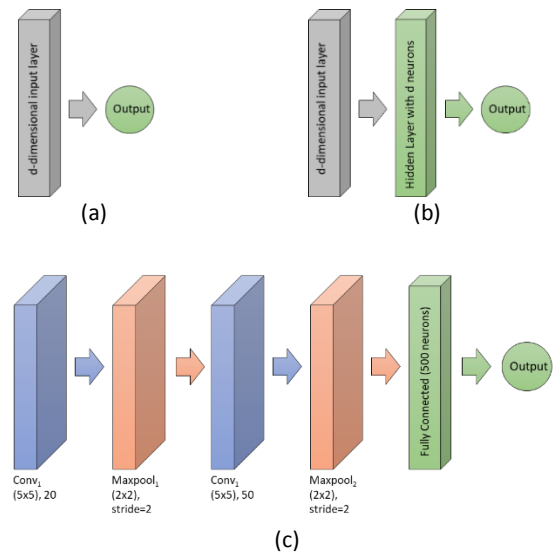

**Figure 3- Neural Network Architectures employed for the different experiments. (a) single layer neural network for benchmark evaluation. (b) 1-hidden layer network for benchmark evaluation. (c) CNN architecture for evaluation over MNIST MIL dataset.**

**Table 1- Percentage Accuracy values with standard deviation for different methods over benchmark MIL datasets.**

| Method | Musk-1 | Musk-2 | Fox | Tiger | Elephant |
|---|---|---|---|---|---|
| **mi-SVM** [19] | 87.4 | 83.6 | 58.2 | 78.4 | 82.2 |
| **MI-SVM** [19] | 77.9 | 84.3 | 57.8 | 84.0 | 84.3 |
| **MI-Kernel** [18] | 88.0 ± 3.1 | 89.3 ± 1.5 | 60.3 ± 2.8 | 84.2 ± 1.0 | 84.3 ± 1.6 |
| **EM-DD** [17] | 84.9 ± 4.4 | 86.9 ± 4.8 | 60.9 ± 4.5 | 73.0 ± 4.3 | 77.1 ± 4.3 |
| **mi-Graph** [20] | 88.9 ± 3.3 | 90.3 ± 3.9 | 62.0 ± 4.4 | 86.0 ± 3.7 | 86.9 ± 3.5 |
| **miVLAD** [21] | 87.1 ± 4.3 | 87.2 ± 4.2 | 62.0 ± 4.4 | 81.1 ± 3.9 | 85.0 ± 3.6 |
| **miFV** [21] | **90.9 ± 4.0** | 88.4 ± 4.2 | 62.1 ± 4.9 | 81.3 ± 3.7 | 85.2 ± 3.6 |
| **mi-Net** [34] | 88.9 ± 03.9 | 85.8 ± 4.9 | 61.3 ± 3.5 | 82.4 ± 3.4 | 85.8 ± 3.7 |
| **MI-Net** [34] | 88.7 ± 4.1 | 85.9 ± 4.6 | 62.2 ± 3.8 | 83.0 ± 3.2 | 86.2 ± 3.4 |
| **MI-Net with DS** [34] | 89.4 ± 4.2 | 87.4 ± 4.3 | 63.0 ± 3.7 | 84.5 ± 3.9 | 87.2 ± 3.2 |
| **MI-Net with RC** [34] | 89.8 ± 4.3 | 87.3 ± 4.4 | 61.9 ± 4.7 | 83.6 ± 3.7 | 85.7 ± 4.0 |
| **Attention** [35] | 89.2 ± 4.0 | 85.8 ± 4.8 | 61.5 ± 4.3 | 83.9 ± 2.2 | 86.8 ± 2.2 |
| **Gated-Attention** [35] | 90.0 ± 5.0 | 86.3 ± 4.2 | 60.3 ± 2.9 | 84.5 ± 1.8 | 85.7 ± 2.7 |
| **Previous Best Performance** | **90.9 ± 4.0** (miFV) | 90.3 ± 3.9 (mi-Graph) | 63.0 ± 3.7 (MI-Net DS) | 86.0 ± 3.7 (mi-Graph) | 86.9 ± 3.5 (mi-Graph) |
| **Proposed Model- Single Layer** | 89.6 ± 1.3 | **90.6 ± 0.4** | **65.8 ± 1.3** | 86.5 ± 1.5 | 83.2 ± 1.5 |
| **Proposed Model- 1 Hidden Layer** | 89.8 ± 0.9 | 89.3 ± 0.4 | 65.5 ± 0.8 | **88.5 ± 1.2** | **87.1 ± 1.3** |

For MNIST MIL, a Convolution Neural Network (CNN) consisting of two convolutional layers and two fully connected layers was used. We used a similar architecture to the one used by Ilse et al. [35] except that we have removed the attention block used by them. The first convolutional layer had a kernel size of 5×5 and output channel size of 20. Rectified Linear Unit (ReLU) activation is applied to the output of this layer.

The next layer performs max-pooling with kernel size of 2×2 and stride of 2. The output of this layer is fed to the next convolutional layer that uses a 5×5 kernel and has 50 output channels. ReLU activation is applied to the output of this layer as well. A max-pooling layer with kernel size 2×2 and stride of 2 follows the convolutional layer. Next is a fully-connected, ReLU activated layer comprising of 500 neurons, which is further connected to the last layer that consists of a single neuron with linear activation. The complete architecture is illustrated in figure 2c.

*2.2.3. Evaluation Protocol and Performance Metrics*

We have compared the performance of our method with existing MIL models: mi-SVM, MI-SVM [19], MI-Kernel [18], EM-DD [17], mi-Graph [20], miVLAD, miFV [21], mi-Net and its variants [34], as well as Attention and Gated Attention Networks [35]. For benchmark datasets, i.e., MUSK-1/2, Fox, Tiger and Elephant, we have used 5 runs of 10-fold cross-validation and percentage bag accuracy as the performance metric for a fair comparison with existing techniques, as the same protocol and performance metric has been used in previous works. Mean and standard deviation of accuracy over 5 runs is reported.

For Multiple Instance MNIST dataset, we have separate train and test sets sampled from the original MNIST dataset as described in the previous section. In line with the work by Ilse et al. [35], bag level AUC-ROC [40] is used as the performance evaluation metric. Test performance averaged over 5 runs for varying bag sizes and training set sizes is reported. Performance comparison with the attention based methods has been presented [35].

**Table 2- Percentage AUC-ROC scores for MNIST based MIL dataset for mean bag length of 10 examples per bag**

| | No. of Training Bags | | | | | | |
|---|---|---|---|---|---|---|---|
| **Methods** | **50** | **100** | **150** | **200** | **300** | **400** | **500** |
| Attention [35] | 76.8 ± 5.4 | 94.8 ± 0.7 | 94.9 ± 0.6 | 97.0 ± 0.3 | 98.0 ± 0.0 | 98.2 ± 0.1 | 98.3 ± 0.2 |
| Gated Attention [35] | 75.3 ± 5.4 | 91.6 ± 1.3 | 95.5 ± 0.3 | 97.4 ± 0.2 | 98.0 ± 0.4 | 98.3 ± 0.2 | 98.7 ± 0.1 |
| **Proposed Method** | 87.6 ± 3.6 | 94.4 ± 2.3 | 95.3 ± 0.8 | 97.0 ± 0.8 | 97.9 ± 0.2 | 98.2 ± 0.2 | 98.5 ± 0.1 |

## 3. Results and Discussion

In this section we present results compiled over the experiments described in the previous section.

### 3.1. Benchmark Datasets

Accuracy values over benchmark datasets: Musk-1/2, Fox, Tiger and Elephant are presented in Table 1. A comparison with other methods is also given. It can be seen that our method performs as well or better than other more complicated neural network based methods: mi-Net, MI-Net and Attention networks [34], [35]. We have presented the comparison with the previous best performing method in the literature. It can be seen that no single method gives the highest performance for all datasets. The highest accuracy for Musk-1 dataset has been reported by miFV [21] as 90.9% with a standard deviation of 4.0%. Our method with one hidden layer produces a comparable 89.8% accuracy with a much lower standard deviation of 0.9%. mi-Graph [20] was previously the best performing method for Musk-2, Tiger and Elephant datasets with percentage accuracies of 90.3%, 86% and 89% respectively. Our method outperforms it in all the three cases with 90.6%, 88.5% and 87.1% accuracy, respectively. Although the improvement in mean accuracy for Musk-2 and Elephant datasets is not large, the standard deviation of accuracy for our method is significantly better. For Elephant dataset the previous best accuracy of 63.0% with a standard deviation of 3.7% was reported for MI-Net with DS (Deep Supervision) [34]. A single layer neural network trained using our proposed scheme produced an improved 65.8% accuracy with a significantly lower standard deviation of 1.3%. Our method shows consistently good performance over all benchmark datasets.

### 3.2. MNIST MIL Dataset

The experiments of 9 vs non-9-containing bags generated from MNIST dataset were conducted to study the effectiveness of our proposed loss in convolutional neural networks. As mentioned earlier, we have used the same experimental setup proposed by Ilse et al. [35] for evaluation of their proposed attention networks based scheme. The percentage AUC-ROC scores computed over a test set of 1000 bags using training sets of varying sizes are given in table 2. We present a comparison with Attention and Gated Attention networks based solution [35].

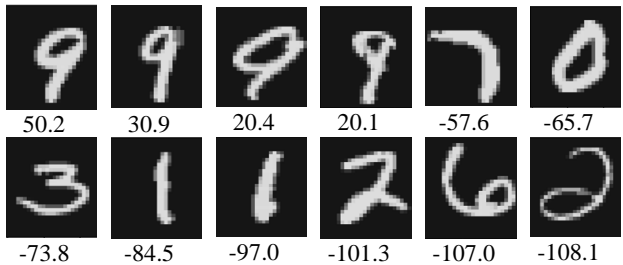

50.2  30.9  20.4  20.1  -57.6  -65.7
-73.8  -84.5  -97.0  -101.3  -107.0  -108.1

**Figure 4- Scores for a positive test bag examples produced by a model trained using the proposed scheme. It can be seen that higher scores are produced for 9 images as compared to non-9s.**

It can be seen that for a bag size as small as 10 and smaller number of training instances (50), our method performs considerably better. Our method produces AUC-ROC of 87.8% with a standard deviation of 3.6% for 50 training bags in comparison to 76.8% produced by Attention Networks. This behavior can be attributed to the use of ranking-like loss function, which, being a paired input loss increases the effective dataset size employed for training, and hence a better generalization performance is seen even for small training dataset. For larger training set sizes, our method produces comparable results.

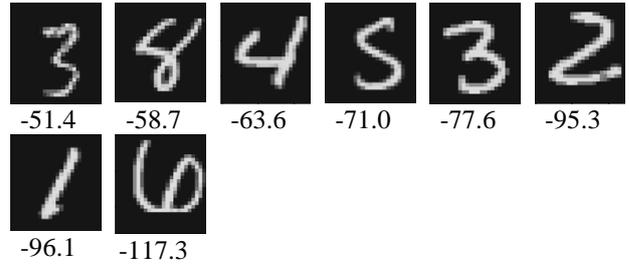

-51.4  -58.7  -63.6  -71.0  -77.6  -95.3
-96.1  -117.3

**Figure 5- Scores for a negative test bag examples produced by a model trained using the proposed scheme.**

To further analyze the trained model, we studied the scores generated for positive and negative examples for a bag. The loss used for training constrains the model to produce higher bag scores for positive bags as compared to the negative ones. We define bag score as the highest score produced by any example in the bag. Scores generated for a positive test bag by a model trained over 500 training bags with 10 examples each on average are shown in figure 4. It can be seen that the model produces higher scores for 9s as compared to non-9s in a bag. This shows that example-level classification can also be performed effectively using the proposed method. To further prove our point, we present the scores generated by the same model for a negative bag in figure 5. It can be seen that the highest score produced by the negative bag is smaller than the one produced by the positive bag.

**Conclusion**

In this paper, we have presented a simplified approach to Multiple Instance Learning using neural networks. We have proposed a ranking like loss function that forces a neural network to produce higher scores for positive bags as compared to the negative ones. Our method is simpler to comprehend and implement as it does not involve any specialized layers and connections to perform Multiple Instance Classification, rather it is based on a simple change in loss function. We have proved the effectiveness of the method on 5 benchmark MIL datasets containing pre-computed handcrafted features. In addition, we have tested the proposed method for CNN based multiple instance learning over a dataset generated from the well-known MNIST data. Results show that, despite being simpler, our approach produces comparable and in some cases better results than other complex methods for neural multiple instance learning. Our method has shown better performance even in cases where training set sizes are small. This property makes the method useful for data-scarce problems as well.

**Acknowledgments**

Amina Asif is funded via Information Technology and Telecommunication Endowment Fund at Pakistan Institute of Engineering and Applied Sciences.

**Conflict of Interest**

The authors declare no conflict of interest.